\documentclass[letterpaper,10pt,conference]{ieeeconf}
\usepackage{amsmath,amssymb,booktabs,cite,graphicx,siunitx,url}
\newcommand{\methodname}{\textsc{Inspect}}
\newcommand{\supstate}{\mathsf{sup}}
\newcommand{\constate}{\mathsf{con}}
\newcommand{\insstate}{\mathsf{ins}}
\title{INSPECT: Learning Robot View Selection from Assistant Use}
\author{Di Wen$^{1}$, Kailun Yang$^{2}$, Wenhao Guo$^{3}$, Yitian Shi$^{1}$, Junwei Zheng$^{1}$, Yufan Chen$^{1}$,\\
Ruiping Liu$^{1}$, Jiale Wei$^{1}$, Rania Rayyes$^{1}$, and Kunyu Peng$^{1,\dagger}$\\[3pt]
{\small $^{1}$Karlsruhe Institute of Technology, Karlsruhe 76131, Germany
\quad $^{2}$Hunan University, Changsha 410012, China}\\
{\small $^{3}$Independent Researcher}\\
{\small First author (email: \texttt{di.wen@kit.edu})
\quad $^{\dagger}$Corresponding author (email: \texttt{kunyu.peng@kit.edu})}}
\begin{document}
\bstctlcite{IEEEexample:BSTcontrol}
\maketitle

\begin{abstract}
Robots inspecting an assembly must determine which parts are present and whether they are correctly installed. During egocentric assembly assistance, head motion and workpiece handling reveal evidence for these checks, while spoken state confirmations link observations to procedural outcomes. We introduce \methodname{}, which learns robot view preferences from records of a smart-glasses assistant that answers part queries and provides next-step guidance. Presence-Invariant TwinSwap (PI-TwinSwap) calibrates object evidence through paired identity interventions. Claim-indexed supervision separates evidence requirements from camera-reproducible observation changes. Object-centered calibration adapts relative view preferences to robot poses, while clause-level screening checks predicted evidence. The robot selects views using only its current observation and known poses, without candidate images. Evaluation uses annotated assistant-video replay to simulate state feedback, without target-domain view labels for policy training. On images of physical gearbox assemblies, \methodname{} achieves the highest view utility among the compared non-oracle policies and raises human-rated full verifiability from 34.8\% to 41.7\% compared with keeping the current view. On commercial angle-grinder recordings in IMPACT, the transferred relative-view selector increases the correct decision rate from 50.6\% to 54.3\% with a frozen perception head. The source code is available at \url{https://github.com/Kratos-Wen/INSPECT}.
\end{abstract}

\section{Introduction}
\label{sec:introduction}

Industrial assembly inspection must establish both part identity and installation state. Recognizing the correct component does not establish whether it is aligned or fully inserted; the relevant evidence depends on the viewpoint and the condition being checked. Active-perception methods address observation choice through geometric utilities or learned view policies~\cite{bajcsy1988active,Breyer2022ClosedLoopNP,jauhri2024actpermoma,koo2025vgavs}. Assembly assistance offers a complementary source of supervision: observations associated with the checks an operator makes while performing the task.

Egocentric assistants support guidance, progress monitoring, and error correction~\cite{wang2023holoassist,flaborea2024prego,lee2024error,wen2025mica,xu2026pro2assist}. Consider an operator using smart glasses to ask about a part or the next assembly step. The assistant checks visual evidence and seeks verbal confirmation when the current step is uncertain. As the operator handles the workpiece and changes head position, new details become visible. Recording these observations together with the state being checked and the user's response connects visual changes to procedural outcomes. \emph{The evidence that grounds assembly guidance can therefore also supervise where a robot should inspect.}

Human-video methods already transfer manipulation and camera behavior~\cite{wang2023mimicplay,jain2024vid2robot,kareer2024egomimic,xiong2025via,lin2026activemimic,sun2026lime}. Learning inspection from assistance requires more than reproducing an observed motion. A claim may become decidable because a viewpoint reveals previously hidden evidence, but it may also be resolved because the operator changes the assembly itself. These events do not provide equivalent supervision for a camera. Transferable changes must further be expressed relative to the workpiece to guide a robot starting from a different pose. The central distinction is between \emph{which evidence resolves a state question} and \emph{which observation changes a robot camera can reproduce}.

\begin{figure*}[t]
\centering
\includegraphics[width=0.90\textwidth]{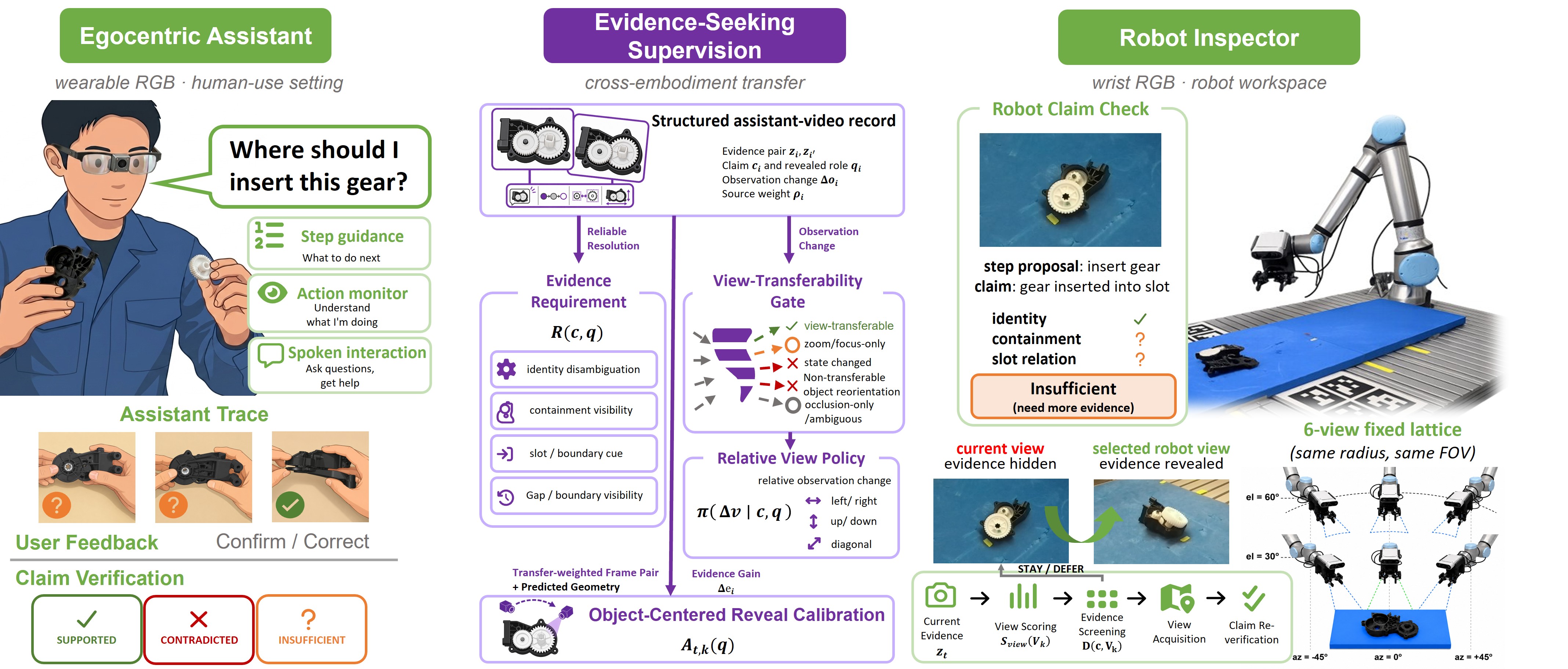}
\caption{\textbf{The evidence that grounds assistance also teaches robots where to inspect.} Claim resolutions identify required evidence; camera-reproducible changes supervise relative view preferences. Object-centered calibration transfers these preferences to robot poses. Selection uses current RGB and known geometry, without candidate images.}
\label{fig:teaser}
\end{figure*}

\methodname{} uses this distinction to connect assembly assistance with robot inspection (Fig.~\ref{fig:teaser}). Records of resolved claims supervise evidence requirements, while camera-reproducible changes supervise relative view preferences. Object-centered calibration connects those preferences to robot poses. The assistant thus supports the operator and supplies supervision for subsequent robot observation selection.

Experiments test this transfer on real assembly objects. Gearbox inspection evaluates access to claim-disambiguating evidence; commercial angle-grinder recordings in IMPACT~\cite{wen2026impact} test observation selection with a different, frozen perception model. Neither uses target-domain view labels to train the policy. Our contributions are:
\begin{itemize}
\item a supervision construction that organizes assembly-assistance records by claims and evidence roles, separating evidence resolution from camera-reproducible observation changes;
\item a relative view policy combining object-centered calibration and clause-level evidence screening for selection without candidate images; and
\item an evaluation on real gearbox assemblies and cross-assembly transfer to commercial angle-grinder recordings, with component and frozen-perception comparisons.
\end{itemize}

\section{Related Work}
\label{sec:related_work}

\subsection{Procedural Assistance and Verification}

Egocentric research spans activity recognition~\cite{damen2022epic,grauman2022ego4d}, procedural activities and execution errors~\cite{sener2022assembly101,schoonbeek2024industreal,grauman2024ego,wen2026impact}, and procedural assistance~\cite{wang2023holoassist,li2025egoproceassist,li2026egoprocevqa}. Error detectors exploit temporal structure, relations, and task graphs~\cite{flaborea2024prego,lee2024error,lee2025gtg,lee2026axg}; proactive assistants use progress estimates to guide intervention~\cite{xu2026pro2assist,kundu2026planwatch}. Snap, Segment, Deploy and MICA integrate task-trained perception with spoken industrial guidance~\cite{wen2025snap,wen2025mica}. We connect claim-level evidence deficits to subsequent observation acquisition.

Selective prediction provides a reject option~\cite{geifman2019selectivenet}, and robot systems use uncertainty to request help or detect failures~\cite{ren2023robots,xu2025can,hung2021introspective}. In \methodname{}, abstention identifies an evidence-seeking opportunity; camera-reproducible resolutions supply supervision for obtaining missing evidence.

\subsection{Learning from Human Observations}

Human-video methods transfer task intent or manipulation behavior~\cite{wang2023mimicplay,jain2024vid2robot,kareer2024egomimic}; human interaction images also supervise functional grasp contacts~\cite{yang2025multikeypoint}. See What I See captures first-person assembly viewpoints to present parts for human use~\cite{wang2020seewhat}. EDITH combines smart-glasses observations, gaze, and speech to ground robot actions in human intent~\cite{lee2026edith}. We instead learn where the robot should observe to verify an assembly claim.

ViA and ActiveUMI collect active-perception demonstrations~\cite{xiong2025via,zeng2025activeumi}, and ActiveMimic recovers camera and wrist trajectories~\cite{lin2026activemimic}. LIME mines intents and observation-gain descriptions paired with relative camera poses from egocentric video~\cite{sun2026lime}. Our supervision indexes observation changes by claims checked during assistance and the evidence roles they require. Transfer eligibility identifies which evidence-revealing changes a robot camera can reproduce.

\subsection{Active Perception and View Selection}

Next-best-view methods choose sensing actions for reconstruction, manipulation, or target visibility~\cite{han2022double,Breyer2022ClosedLoopNP,jauhri2024actpermoma,hu2026oa}. VISO-Grasp combines object-centric spatial reasoning, active view planning, and multi-view grasp fusion for occluded targets~\cite{shi2025visograsp}. Query-driven approaches seek observations needed for an answer~\cite{das2018embodiedqa,wang2026avp}. AP-VLM uses language and a spatial grid for exploration, while VAP-TAMP combines vision-language model (VLM) guidance with action knowledge and scene graphs~\cite{sripada2024apvlm,oloo2026vaptamp}. VG-AVS learns next-view selection from the current image~\cite{koo2025vgavs}, sharing our candidate-image restriction. ActiveVLA and SaPaVe couple perception actions with manipulation~\cite{liu2026activevla,liu2026sapave}. Our focus is the supervision source: claim-indexed evidence changes, filtered to separate assembly-state changes from camera-reproducible observations.

\section{Method}
\label{sec:method}

We learn from assistant records where a robot should look to check an assembly claim. Given RGB $o_t$ at pose $V_t$, candidate poses $\mathcal V$, and claim $c$, the robot selects $V_k\in\mathcal V$ without candidate images. It can retain $V_t$ or defer a decision.

For example, recognizing a gear may leave its slot relation unclear. A resolved claim provides supervision for the missing role, while an eligible observation change teaches a relative motion preference. We adapt that preference to robot poses and check the predicted evidence before acquiring a view. Throughout, $t$, $i$, and $k$ index observations, records, and candidates, respectively; $\epsilon>0$ stabilizes computation.

\begin{figure*}[t]
\centering
\includegraphics[width=\textwidth]{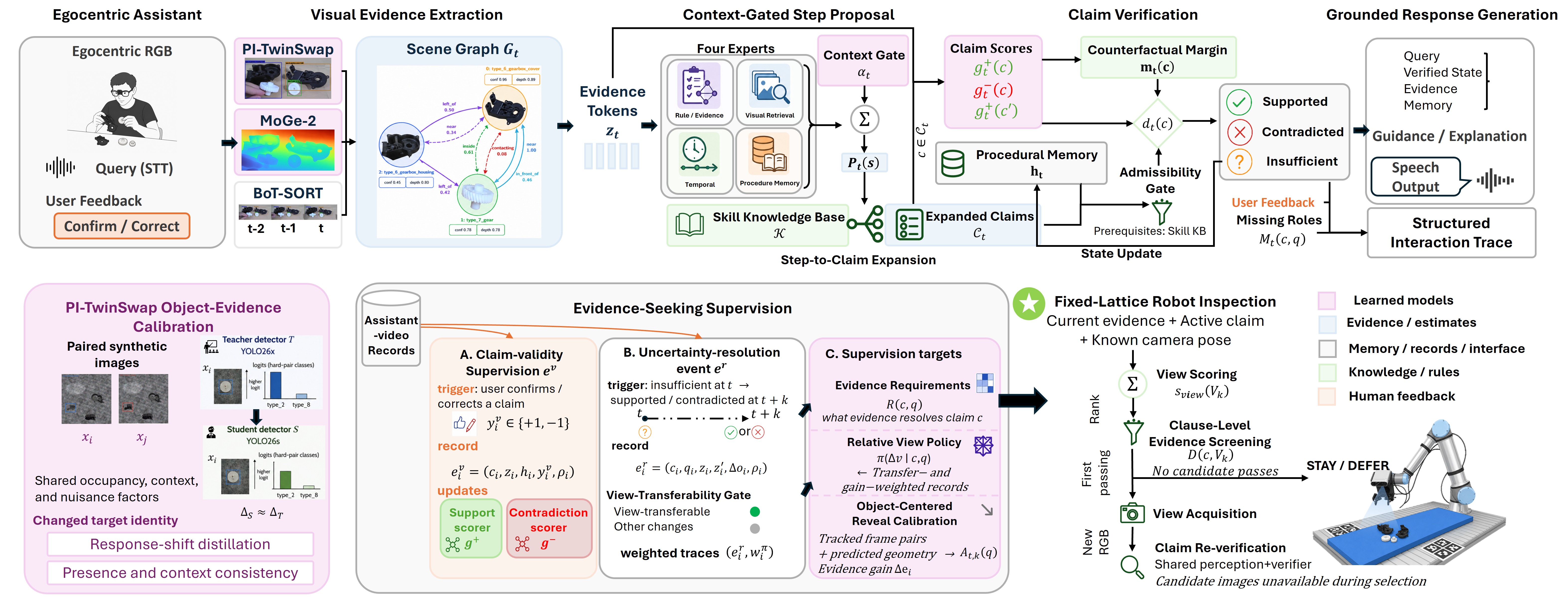}
\caption{\textbf{From assistance to robot inspection.} Calibrated evidence supports claim verification; assistant records supervise evidence requirements and geometrically calibrated view preferences. Selected robot views undergo re-verification.}
\label{fig:method_overview}
\end{figure*}

\subsection{Egocentric Assistant}

The smart-glasses interface uses speech-to-text (STT) for part queries, next-step requests, and state feedback (Fig.~\ref{fig:method_overview}). Confirmations and corrections update fusion and memory, and are logged alongside RGB observations of head motion and workpiece handling.

The task knowledge base (KB) $\mathcal{K}$ specifies steps, claim templates, preconditions, postconditions, and counterfactuals. Claims $c\in\mathcal C$ describe identity or installation state; evidence roles $q\in\mathcal Q$ specify distinguishing cues. The role vocabulary and entity bindings are task-specific inputs to shared verification and observation-learning rules.

We use the same evidence interface for assistance and inspection. Detections, appearance, estimated geometry, relations in scene graph $G_t$, and tracks form $z_t=\phi(o_t)$ from RGB $o_t$, while memory $h_t$ stores supported, contradicted, and pending claims.

The verifier supplies claim state $d_t(c)\in\{\supstate,\constate,\insstate\}$, meaning supported, contradicted, or insufficient, together with missing-role weights $M_t(c,q)$. When neither support nor contradiction can be established, these weights identify the evidence needed for further inspection.

\noindent\textbf{PI-TwinSwap Object-Evidence Calibration.}
\label{sec:pi_twinswap}
We train the detector to distinguish confusable parts under matched visual context. Presence-Invariant TwinSwap (PI-TwinSwap) calibrates a compact student against a teacher using paired identity interventions (Fig.~\ref{fig:method_overview}, lower left). For nuisance configuration $B$, unchanged context objects $\mathcal{O}_{\mathrm{ctx}}$, compositor slot $b$, and confusable identities $p_i,p_j$, the compositor produces
\begin{equation}
\begin{aligned}
x_i&=\Psi(B,\mathcal{O}_{\mathrm{ctx}},b\leftarrow p_i),\\
x_j&=\Psi(B,\mathcal{O}_{\mathrm{ctx}},b\leftarrow p_j).
\end{aligned}
\label{eq:pi_pair}
\end{equation}
The images differ in target identity while sharing slot occupancy, placement, scale, occlusion configuration, and photometric factors. We apply nuisance factors jointly during composition and disable independent image augmentation to preserve this correspondence.

Let $u_D(x,b)$ be slot-local class logits for teacher or student $D\in\{T,S\}$. Knowledge distillation (KD) uses temperature $T_{\mathrm{KD}}>0$; responses are $r_D(x,b)=\operatorname{softmax}(u_D(x,b)/T_{\mathrm{KD}})$ and $\Delta_D=r_D(x_i,b)-r_D(x_j,b)$. With temperature factors absorbed into loss weights, training uses
\begin{equation}
\begin{aligned}
\mathcal{L}_{\mathrm{detector}}
&=\mathcal{L}_{\mathrm{det}}+\lambda_{\mathrm{KD}}\mathcal{L}_{\mathrm{KD}}\\
&\quad+\lambda_{\mathrm{swap}}\mathcal{L}_{\mathrm{swap}}
+\lambda_{\mathrm{hp}}\mathcal{L}_{\mathrm{hp}}+\mathcal{L}_{\mathrm{PI}},\\
\mathcal{L}_{\mathrm{swap}}
&=\operatorname{SmoothL1}(\Delta_S,\Delta_T).
\end{aligned}
\label{eq:pi_twinswap_loss}
\end{equation}
The supervised loss $\mathcal{L}_{\mathrm{det}}$ is augmented by KD and hard-pair terms with weights $\lambda$. $\mathcal{L}_{\mathrm{KD}}$ matches response shifts and identity margins, and $\mathcal{L}_{\mathrm{hp}}$ separates confusable identities. The weighted terms $\mathcal{L}_{\mathrm{PI}}$ align paired presence-energy changes and levels with the teacher while maintaining stable context responses. Presence energy is the log-sum-exp of class logits. Boxes, confidence, and identity margins enter $z_t$.

\noindent\textbf{Context-Gated Step Proposal.}
Step proposal narrows the claims to inspect. Expert $x$ scores step $s$ by $\ell_{t,x}(s)$ using rules and evidence, visual retrieval, temporal continuity, or procedural memory. The context gate $g_\eta$, with parameters $\eta$ and context $\gamma_t$, produces expert weights $\alpha_t$, step probabilities $P_t$, and active claims $\mathcal C_t$:
\begin{equation}
\begin{aligned}
\alpha_t&=\operatorname{softmax}(g_\eta(\gamma_t)),\\
P_t(s)&=\operatorname{softmax}_s\!\left(\sum_x\alpha_{t,x}\ell_{t,x}(s)\right),\\
\mathcal{C}_t&=\operatorname{expand}(\operatorname{TopK}P_t;\mathcal{K}).
\end{aligned}
\label{eq:proposal_to_claim}
\end{equation}
TopK retains $K$ proposals, whose KB claims are instantiated by $\operatorname{expand}$. Interaction and contact cues enter evidence and context. We verify the expanded claims before updating completion history.

\noindent\textbf{Claim Verification.}\label{sec:claim_verification} For $c\in\mathcal{C}_t$, the scorer estimates support $g_t^+(c)$ and contradiction $g_t^-(c)$. Specialized counterfactual claims $\mathcal{C}^{-}(c)$ include confusable identities and relation alternatives; otherwise the negated claim is used. Their support defines the counterfactual margin
\begin{equation}
m_t(c)=g_t^+(c)-\max_{c'\in\mathcal{C}^{-}(c)}g_t^+(c').
\label{eq:counterfactual_margin}
\end{equation}
A contradicted prerequisite sets the admissibility gate $a(c,h_t)\in\{0,1\}$ to zero, yielding $\constate$ as a history-based procedural block. Otherwise visual contradiction requires $g_t^-(c)\geq\tau^-$ and $g_t^-(c)\geq g_t^+(c)+\tau^m$. Support requires verified prerequisites, $g_t^+(c)\geq\tau^+$, $g_t^+(c)\geq g_t^-(c)+\tau^m$, and $m_t(c)\geq\tau^m$. Thresholds $\tau^+,\tau^-,\tau^m$ control support, contradiction, and score separation. An unverified prerequisite withholds support. Cases satisfying neither decision remain insufficient.

\noindent\textbf{Grounded Response Generation.} The responder uses the query, evidence, claim states, and memory to answer, guide, or seek state confirmation without overriding verification. Supported claims update history, contradictions block transitions, and insufficient claims remain pending. We record responses, evidence, and feedback in the structured interaction trace, obtaining supervision without asking users to follow camera-motion instructions.

\subsection{Evidence-Seeking Supervision}

We separate feedback about claim validity from evidence about a useful observation change. The interaction trace provides validity records $e_i^v=(c_i,z_i,h_i,y_i^v,\rho_i)$, with confirmation/correction label $y_i^v\in\{+1,-1\}$ and label confidence $\rho_i$. Insufficient-to-resolved transitions yield $e_i^r=(c_i,q_i,z_i,z_i',\Delta o_i,\rho_i)$, recording the revealed role and observation change.

A workpiece reorientation may reveal which evidence resolves a claim even when a robot camera cannot reproduce that change. We therefore update evidence requirements from reliable role resolutions with weights $w_i^R$, excluding true assembly-state changes. The view-transferability gate separately determines which records contribute motion weights $w_i^\pi$.

We estimate the motion source from central and peripheral keypoint flow. At 720-pixel maximum image side, central flow below 8 pixels is weak. Peripheral flow of at least 4.8 pixels with central/peripheral cosine above 0.45 indicates camera motion, conflicting flows indicate mixed motion, and absent peripheral support indicates object-relative motion. Orbit tokens encode eight coarse camera directions. We reject state changes, occlusion-only or zoom/focus explanations, missing tokens, and object-relative, weak, or unknown motion.

For camera motion, we set $T_i$ to $\max(0.05,\rho_i^{\mathrm{motion}})$. Mixed motion receives $\max(0.05,0.55\rho_i^{\mathrm{motion}})$, and rejected records receive $T_i=0$. Motion confidence combines track retention, displacement, and scale consistency. The count weight $w_i^\pi=T_i\rho_i S_i I_i g_i$ includes stability $S_i$ and KB role importance $I_i$. For the recorded evidence-score gain $\Delta_i^{\mathrm{score}}$ (after minus before), we use $g_i=\operatorname{clip}_{[0.05,1]}(\Delta_i^{\mathrm{score}})$. Clipping gives nonpositive gains a floor of 0.05; transfer-ineligible records still have zero weight. This replay sets $\rho_i=\rho_i^{\mathrm{motion}}$ and $S_i=1$. Evidence requirements are
\begin{equation}
R(c,q)=\frac{\beta R_0(c,q)+\sum_i w_i^R\mathbf{1}[c_i=c,q_i=q]}{\beta+\sum_i w_i^R\mathbf{1}[c_i=c]+\epsilon}.
\label{eq:requirement}
\end{equation}
The indicator $\mathbf1[\cdot]$ selects matching records, which update the normalized ontology prior $R_0$ with prior strength $\beta$. To share evidence across related claims, we pool records within a parent relation schema and use its posterior as the prior for type-specific updates.

We group eligible tokens $\Delta v_i\in\Delta\mathcal V$ into left--right-reflected action families, where $\Delta\mathcal V$ is the discrete relative-action space. Transfer- and gain-weighted counts for family $a$ are smoothed through claim--role, claim, role, and global distributions:
\begin{equation}
P(a\mid c,q)=\frac{N_{cq,a}+\alpha P_{\mathrm{parent}}(a\mid c,q)}{\sum_{a'}N_{cq,a'}+\alpha}.
\label{eq:relative_prior}
\end{equation}
Weighted event mass $N_{cq,a}$ competes with prior mass $\alpha>0$ from $P_{\mathrm{parent}}$, so sparsely observed claim--role pairs rely more on broader distributions. The relative view policy $\pi(\Delta v\mid c,q)$ divides family probabilities among reflected members. These coarse preferences are then calibrated geometrically.

\subsection{Object-Centered Reveal Calibration}

To adapt a relative direction to a new robot pose, we align 3D points from temporal keypoints and monocular point maps. Inverting the robust similarity transform places camera centers in the tracked reference frame, where viewing rays give azimuth and elevation changes in radians. After filtering non-reproducible motion, calibration selects a parallax kernel for angular magnitudes or a directional kernel for both angles; $\pi$ supplies discrete directions.

For segment outcome $y_i\in\{\supstate,\constate\}$, let $E_i(t)$ be the corresponding frozen-scorer score. Endpoint gain $\Delta e_i=E_i(t_i')-E_i(t_i)$ defines target $y_i^g=\operatorname{sign}(\Delta e_i)\max(|\Delta e_i|-0.05,0)$. We classify positive targets as helpful and nonpositive targets as unhelpful. Each record's weight multiplies transfer, label-confidence, stability, and KB role-importance factors, divided by the transition's record count. Stability is 1, with lower importance for role fallback. Signed gain is the geometric prediction target, whereas $g_i$ scales discrete action counts.

For claim--role--counterfactual context $(c,q,f)$, a kernel weights records by motion similarity. Let $\mathrm{dist}_i^2$ be squared bandwidth-normalized distance, $s_i$ semantic compatibility, $r_i$ reference-frame compatibility, and $Q,Q_i$ query and record quality. The weight is
\begin{equation}
k_i=\exp(-\mathrm{dist}_i^2/2)s_i r_i
\sqrt{\max(0.05,Q)\max(0.05,Q_i)}.
\label{eq:record_kernel}
\end{equation}
Let $n^+$ and $n^-$ denote weighted helpful and unhelpful outcomes, $u$ the weighted signed-gain sum, and $n$ the observation mass. The geometry-conditioned helpful probability $p_g$ and expected evidence gain $\mu_g$ shrink toward context-only estimates $p_0,\mu_0$:
\begin{equation}
p_g=\frac{n^++\lambda_K p_0}{n^++n^-+\lambda_K},
\qquad
\mu_g=\frac{u+\lambda_K\mu_0}{n+\lambda_K}.
\label{eq:geometry_prediction}
\end{equation}
With prior strength $\lambda_K$, weakly supported predictions approach the context estimates $p_0,\mu_0$ stored in geometric memory. We use the departure from these estimates to measure what motion geometry adds.

For a candidate view, let $\omega_f$ be normalized counterfactual-category weights. Write $\Delta\ell_f=\operatorname{logit}p_{g,f}-\operatorname{logit}p_{0,f}$ for the geometric log-odds correction. The frozen one-step selector computes
\begin{equation}
\begin{aligned}
\delta_{t,k}(q)
&=\sum_f\omega_f\Bigl[\Delta\ell_f
+\lambda_g\frac{\mu_{g,f}-\mu_{0,f}}{s_g}\Bigr],\\
A_{t,k}(q)
&=\operatorname{clip}_{[0.25,4]}\!
\Bigl(\exp\Bigl[\lambda_A\frac{n_{t,k}}{n_{t,k}+s_0}\delta_{t,k}(q)\Bigr]\Bigr).
\end{aligned}
\label{eq:geometry_correction}
\end{equation}
Geometric support $n_{t,k}=\sum_f\omega_f n_f$ controls correction strength: scale $s_0$ keeps it near neutral when few records support a candidate. Gain scale $s_g$ and coefficients $\lambda_A,\lambda_g$ control calibration. Probabilities are clipped to $[10^{-5},1-10^{-5}]$ before logits. Reference weight $r_i$ is 1 for compatible types and a calibrated penalty otherwise. Tracking quality combines inlier and visibility fractions with exponential penalties for normalized fit residual and cycle error. Calibrated robot poses use $Q=1$.

\subsection{Fixed-Lattice Robot Inspection}

We now combine the learned preferences with evidence from the current robot view. RGB at pose $V_t$ supplies $z_t$ and role scores $e_{tq}\in[0,1]$ for claim $c$, with missing evidence $M_t(c,q)=\operatorname{clip}_{[0,1]}(1-e_{tq}/0.65)$. Identity uses confidence and margin, while spatial roles use relation and boundary cues. Candidate poses $\mathcal V$ share radius, intrinsics, and workspace coverage. View scoring uses their displacements $\Delta_{t,k}=\Delta(V_t,V_k)$:
\begin{equation}
\begin{aligned}
B_t(c,q)&=R(c,q)M_t(c,q)\bar H_t(c,q),\\
G_{t,k}(q)&=A_{t,k}(q)\sum_{\Delta v}\pi(\Delta v\mid c,q)
\kappa(\Delta v,\Delta_{t,k}),\\
S_{\mathrm{view}}(V_k)&=\sum_q B_t(c,q)G_{t,k}(q)-\lambda_d d(V_t,V_k).
\end{aligned}
\label{eq:view_score}
\end{equation}
$B_t$ weights missing roles and $G_{t,k}$ their predicted reveal benefit; $\lambda_d$ weights motion cost $d$. Let $\chi(f,q)$ measure compatibility between counterfactual category $f$ and role $q$ in the KB, and $J_t(c,q)=\sum_f\omega_f \chi(f,q)$. Known category $f_t$ gives $H_t=\chi(f_t,q)J_t$; otherwise $H_t=J_t$. Normalization uses $\bar H_t=H_t/Z_t$, with $Z_t$ the requirement-weighted mean over missing roles (one if zero). For azimuth/elevation changes $(\Delta\psi,\Delta\theta)$, $\kappa$ is fourth-power nonnegative cosine similarity with $(\Delta\psi/45^\circ,\Delta\theta/30^\circ)$; $d=|\Delta\psi|/45^\circ+|\Delta\theta|/30^\circ$.

Improving one evidence role can leave another unresolved. We therefore screen candidate views against the whole claim. With $\mathcal H$ denoting entropy, let $v_{t,q}$ be threshold-normalized current evidence clipped to $[0,1]$, and $\zeta_{cq}=1-\mathcal{H}(\pi(\cdot\mid c,q))/\log|\Delta\mathcal{V}|$ the action concentration. Define $L_{kq}=\sqrt{A_{t,k}(q)}\sum_{\Delta v}\pi\kappa$, using the arguments in Eq.~\ref{eq:view_score}, and $\widehat L_{kq}=L_{kq}/\max\{\epsilon,\max_\ell L_{\ell q}\}$. Predicted evidence is
\begin{equation}
\begin{aligned}
\nu_{k,q}&=\operatorname{clip}_{[0,1]}
\bigl[v_{t,q}+\Delta\nu_{k,q}-\ell_{kq}\bigr],\\
\Delta\nu_{k,q}&=(1-v_{t,q})\sqrt{\zeta_{cq}}J_t(c,q)\widehat L_{kq}.
\end{aligned}
\label{eq:role_projection}
\end{equation}
The preservation loss is $\ell_{kq}=v_{t,q}\max(0,1-F_{kq})$, where $F_{kq}$ is the candidate-to-current role-visibility ratio predicted from current geometry, clipped to $[0.25,4]$; unavailable geometry gives $F_{kq}=1$.

Roles within clause $\mathcal{Q}_j$ offer alternative cues, whereas active clauses encode requirements that must be considered together. We aggregate with a weighted complement within each clause and a weighted geometric mean across clauses. The resulting soft evidence score is not a calibrated probability of logical satisfaction. With $\bar b_j=b_j/\sum_\ell b_\ell$,
\begin{equation}
\begin{aligned}
\sigma_j(c,V_k)&=1-\prod_{q\in\mathcal{Q}_j}
\bigl[\max\{\epsilon_D,1-\nu_{k,q}\}\bigr]^{a_{jq}},\\
D(c,V_k)&=\exp\!\Bigl[\sum_j\bar b_j
\log\max\{\epsilon_D,\sigma_j(c,V_k)\}\Bigr].
\end{aligned}
\label{eq:decidability}
\end{equation}
Nonnegative weights $a_{jq}$ sum to one within clause $j$, $b_j$ weights clauses, and $\epsilon_D>0$ stabilizes computation. We rank candidates by $S_{\mathrm{view}}$ and acquire the first satisfying $S_{\mathrm{view}}(V_k)\geq\tau^V$ and $D(c,V_k)-D(c,V_t)-\lambda_d d(V_t,V_k)>\tau^V$. Both cutoffs are fixed to $\tau^V$ as an implementation choice, despite their different score scales. If no candidate qualifies, the robot stays or defers. The shared verifier evaluates the acquired image before committing a claim.

\section{Experiments}
\label{sec:experiments}

\subsection{Setup}

We first evaluate view selection and its transfer mechanism, then examine the supporting verifier and detector. Inspection experiments use offline one-step replay of recorded observations.

\noindent\textbf{Implementation.} YOLO26s is initialized from Gear8 training in Snap, Segment, Deploy~\cite{wen2025snap}; PI-TwinSwap uses a YOLO26x teacher~\cite{jocher2026yolo26}. Geometry and tracks use MoGe-2~\cite{wang2025moge2} and BoT-SORT~\cite{aharon2022bot}. The gearbox KB binds eight roles: identity disambiguation, insertion, containment, slot relation, gap visibility, boundary alignment, contact, and claim disambiguation. The last separates claim alternatives; occlusion is a quality cue. Small- and large-gear insertion share an identity-and-spatial parent schema with part-specific records; cover seating uses gap, boundary, and contact roles.

\noindent\textbf{Supervision and Selection.} We replay video annotations to simulate spoken state confirmations and corrections reproducibly. Annotations provide claim identities, outcomes, intervals, and resolution reasons, while evidence and motion are predicted and the KB assigns roles. Requirements use 115 role-resolution records from 25 episodes in 15 videos. Discrete supervision has 26 motion-role records from 14 transitions in seven videos. Geometric supervision projects 151 frame pairs into 453 role records, retaining 47 from 14 videos. These pools overlap: each transition or frame pair can yield multiple roles. Assistant-video leave-one-video-out calibration selects requirement settings by episode-balanced role negative log-likelihood and geometric settings by helpful-outcome Brier loss. Settings are then frozen, with robot and IMPACT view labels excluded from policy training and selection.

Semantic weights $s_i$ are 1 for a matched role--counterfactual pair, 0.35 for either match, and 0.08 otherwise, multiplied by 1.15 for the same claim; the pooled fallback uses 1 before the claim multiplier. The geometric correction uses $\lambda_A=\lambda_g=1$, $s_0=2$, and $s_g\approx0.109$. The frozen parallax kernel has bandwidth 0.75 rad and reference-mismatch weight 0.15; the identity kernel uses directional bandwidths $(0.45,0.55)$ rad with prior strength $10^6$, making its geometric correction negligible relative to the context prior. The robot uses $\lambda_d=0.05$ and $\tau^V=0.02$, without policy updates during one-step evaluation.

\noindent\textbf{Gearbox Inspection.} Of 60 physical setups with 360 images, we evaluate all 22 setups assigned steps 2--4 from six start views each; the remaining 38 are step 1. We also aggregate paired gains per setup because the 132 trials share scenes. Policies share the frozen trigger, observations, and costs. Claims and assembly families are supplied, while current-image evidence and geometry are predicted. Two annotators blinded to model outputs graded view utility as non-informative (0), partially informative (1), or decidable from that image alone (2). These labels score completed selections. Every evaluated setup has a grade-2 view, although eligibility uses task stage alone. We report mean utility and Verifiable@1, the fraction graded 2, separately from verifier commitment.

\noindent\textbf{Baselines.} Current View stays, Uniform Non-current View averages the five alternatives when triggered, and Minimum-Cost View averages minimum-cost ties. The claim-conditioned prior omits relative evidence-change supervision. Qwen3-VL-4B-Instruct~\cite{bai2025qwen3vl} (Qwen3-VL in tables) receives current RGB, the claim, current-view ID, and lattice geometry. The contextual variant adds procedure text and the predicted scene graph. Both use the same frozen checkpoint, deterministic decoding, inspection actions, and candidate-image restriction.

\subsection{Assistant-to-Robot View Selection}

In Table~\ref{tab:inspection}, uninformed motion leaves mean utility essentially unchanged, whereas claim conditioning helps. \methodname{} achieves the highest utility and full verifiability among the compared non-oracle policies, improving utility in 15 of 22 setups, tying in three, and decreasing in four. Qwen3-VL improves average utility without increasing full verifiability, illustrating why we measure both evidence gain and claim decidability.

\begin{table}[t]
\centering
\caption{Fixed-lattice view selection on 132 matched trials with candidate images hidden. Utility is human-rated; Verifiable@1 counts grade-2 views. Qwen3-VL~\cite{bai2025qwen3vl} uses the 4B-Instruct checkpoint; $^\dagger$ marks the oracle upper bound.}
\label{tab:inspection}
\setlength{\tabcolsep}{0.35em}
\resizebox{\ifdim\width>\columnwidth\columnwidth\else\width\fi}{!}{%
\begin{tabular}{@{}l *{2}{S[table-format=1.3]}@{}}
\toprule
Policy & {Utility $\uparrow$} & {Verifiable@1 $\uparrow$}\\
\midrule
Current View & 1.045 & 0.348\\
\addlinespace[0.25em]
Uniform Non-current View & 1.035 & 0.350\\
\addlinespace[0.25em]
Minimum-Cost View & 1.043 & 0.343\\
\addlinespace[0.25em]
Qwen3-VL (RGB + claim) & 1.076 & 0.333\\
\addlinespace[0.25em]
Qwen3-VL (+ procedure + scene graph) & 1.076 & 0.326\\
\addlinespace[0.25em]
Claim-conditioned view prior & 1.098 & 0.356\\
\addlinespace[0.25em]
\methodname{} & \bfseries 1.205 & \bfseries 0.417\\
\midrule
Oracle View$^\dagger$ & 2.000 & 1.000\\
\bottomrule
\end{tabular}%
}
\end{table}

On gearbox replay, improved human-rated verifiability did not yield additional correct commitments after view changes.

\subsection{Transfer Mechanism Analysis}

Table~\ref{tab:components} examines which parts of transfer account for the gains. Admitting all 453 records (Without geometric transfer filtering) lowers utility at fixed kernel settings and discrete policy. More records therefore do not compensate for ineligible changes here. Removing object-centered calibration also lowers utility, supporting the adaptation of coarse preferences to reference geometry under the fixed settings.

Without clause screening, expected-gain ranking yields fewer fully verifiable views. Requirement weighting has a different outcome: removing it leaves both metrics unchanged, so its independent benefit is not established by this ablation.

Conditional-count controls additionally test whether a global motion preference suffices. Removing claim and claim--role counts reduces utility to 1.174 and Verifiable@1 to 0.394, with other parameters fixed. The effect is localized to two setups; most final selections remain unchanged.

\begin{table}[t]
\centering
\caption{Selector ablations on the same 132 trials, with all other parameters frozen.}
\label{tab:components}
\setlength{\tabcolsep}{0.35em}
\resizebox{\ifdim\width>\columnwidth\columnwidth\else\width\fi}{!}{%
\begin{tabular}{@{}lcc@{}}
\toprule
Variant & {Utility $\uparrow$} & {Verifiable@1 $\uparrow$}\\
\midrule
Without geometric transfer filtering & 1.136 & 0.379\\
\addlinespace[0.25em]
Without object-centered calibration & 1.152 & 0.402\\
\addlinespace[0.25em]
Without clause-level screening & 1.189 & 0.394\\
\addlinespace[0.25em]
Without evidence-requirement weighting & 1.205 & 0.417\\
\addlinespace[0.25em]
\methodname{} & 1.205 & 0.417\\
\bottomrule
\end{tabular}%
}
\end{table}

\begin{figure}[t]
\centering
\includegraphics[width=\columnwidth]{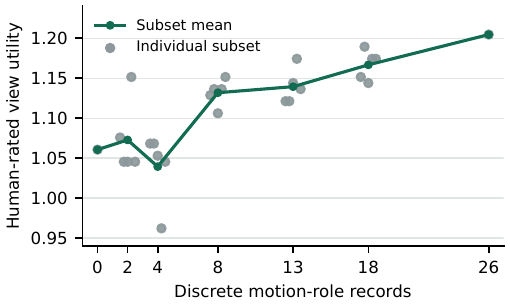}
\caption{\textbf{Sensitivity to discrete motion supervision.} Dots show record subsets with slight horizontal offsets; the line shows mean utility. Requirements and geometry are fixed. Zero removes discrete records; 26 reproduces the frozen selector.}
\label{fig:event_budget}
\end{figure}

\noindent\textbf{Relative-Policy Supervision.} Figure~\ref{fig:event_budget} varies motion-role records at fixed requirements, geometry, and hyperparameters. Records can share transitions. Five subset orderings cover intermediate budgets, with one distinct run per endpoint. All tested subsets containing at least eight records outperform the empty-record baseline, while smaller budgets are non-monotone.

\subsection{Cross-Dataset Transfer}

We extend IMPACT-v1.1's front-view assembly state recognition \texttt{split1}~\cite{wen2026impact} to four static cameras. Hashing (seed 2026) selects one component for each of 267 transitions, each evaluated from four start views. Queries use the official state 45 front-view frames after each transition, with normalized-time alignment across cameras. We train the state head on official training data and VideoMAE V2 features~\cite{wang2023videomaev2}, selecting its checkpoint and cutoff on validation. The fixed adapter removes component-specific identifiers from claims while retaining installation predicates. It maps generic identity evidence to identity disambiguation, alignment to slot relation, containment to insertion verification, and boundary visibility to gap visibility; empty role lists use claim disambiguation.

Frozen-head utility is 2 for a correct non-abstaining decision, 1 for a correct raw decision withheld by the cutoff, and 0 for an incorrect raw decision. Correct Decision Rate counts grade-2 predictions over all trials. Inspection is triggered only when the current prediction abstains. Outcome labels enter scoring only; Table~\ref{tab:impact} separates the selectors.

The relative-view selector ranks learned view scores, whereas the clause-screened selector additionally uses geometric role-visibility factors and checks predicted claim evidence (Table~\ref{tab:impact}). They reach Correct Decision Rates of 54.3\% and 52.8\%, respectively, compared with 50.6\% for Current View, with moves in 7.7\% and 4.4\% of trials. The relative-view selector's 82 inspections yield 40 correct commitments, 29 incorrect commitments, and 13 continued abstentions. Clause-level screening instead improves human-rated verifiability in the gearbox ablation.

\begin{table}[t]
\centering
\caption{Transfer to IMPACT~\cite{wen2026impact} on 1,068 trials. Utility is derived from the frozen state head; Correct Decision Rate counts correct non-abstaining predictions over all trials.}
\label{tab:impact}
\setlength{\tabcolsep}{0.35em}
\resizebox{\ifdim\width>\columnwidth\columnwidth\else\width\fi}{!}{%
\begin{tabular}{@{}l *{2}{S[table-format=1.3]}@{}}
\toprule
Policy & {Utility $\uparrow$} & {Correct Decision Rate $\uparrow$}\\
\midrule
Current View & 1.043 & 0.506\\
\addlinespace[0.25em]
\methodname{} (relative-view selector) & \bfseries 1.098 & \bfseries 0.543\\
\addlinespace[0.25em]
\methodname{} (clause-screened selector) & 1.075 & 0.528\\
\bottomrule
\end{tabular}%
}
\end{table}

\subsection{Supporting Verification and Perception}

To isolate visual verification, we supply annotated active claims and assembly families for 1,315 states from 37 videos in nine scenario groups. Five scenario-grouped outer folds provide predictions, with operating points selected on three grouped inner folds. Pooled outer-fold metrics therefore measure verification conditional on the supplied claim.

\noindent\textbf{Visual Verification Protocol.} Given an annotated active claim, the visual verifier predicts three-state posteriors $p_t$. Its optional past-only filter is $\bar p_t=\xi\bar p_{t-1}+(1-\xi)p_t$, initialized by the first observation of each claim in each video. With contradiction checked first, a decision $d\in\{\supstate,\constate\}$ requires
\begin{equation}
\bar p_t^d\geq\vartheta_d,\qquad
\bar p_t^d\geq\max_{d'\neq d}\bar p_t^{d'}+\delta.
\label{eq:conditional_rule}
\end{equation}
The smoothing coefficient $\xi$ controls the influence of earlier observations, with $\xi=0$ giving frame-wise decisions. Threshold $\vartheta_d$ and posterior margin $\delta$ determine when the verifier abstains. This diagnostic excludes interactive memory gates and the counterfactual claim margin.

Table~\ref{tab:verification} reports support precision (Sup. P.) and recall (Sup. R.); False Support is the fraction predicted supported among contradicted or insufficient states. Macro-F1 averages F1 over the three states. Filtering reduces acceptance of contradicted claims while preserving insufficient-state recognition, selecting a higher-precision operating point at lower supported-claim recall.

\begin{table}[t]
\centering
\caption{Given annotated active claims, (a) tests past-only temporal filtering and (b) compares matched single-frame queries with Qwen3-VL~\cite{bai2025qwen3vl}. $N$ counts claim states; Sup. P. and Sup. R. denote support precision and recall. Neither panel uses interactive history gates.}
\label{tab:verification}
\setlength{\tabcolsep}{0.35em}
\resizebox{\ifdim\width>\columnwidth\columnwidth\else\width\fi}{!}{%
\begin{tabular}{@{}l S[table-format=4.0] *{4}{S[table-format=1.3]}@{}}
\toprule
Model and context & {$N$} & {Sup. P. $\uparrow$} & {Sup. R. $\uparrow$} & {False Sup. $\downarrow$} & {Macro-F1 $\uparrow$}\\
\midrule
\multicolumn{6}{@{}l}{\textit{(a) Scenario-grouped diagnostic}}\\
Verifier (single-frame) & 1315 & 0.793 & \bfseries 0.422 & 0.053 & \bfseries 0.646\\
\addlinespace[0.25em]
Verifier (temporal filter) & 1315 & \bfseries 0.903 & 0.328 & \bfseries 0.017 & 0.615\\
\midrule
\multicolumn{6}{@{}l}{\textit{(b) Matched-query comparison}}\\
Qwen3-VL (RGB + claim) & 60 & 0.520 & \bfseries 0.650 & 0.300 & 0.427\\
\addlinespace[0.25em]
Qwen3-VL (+ procedure) & 60 & 0.520 & \bfseries 0.650 & 0.300 & 0.435\\
\addlinespace[0.25em]
Verifier (single-frame) & 60 & \bfseries 0.900 & 0.450 & \bfseries 0.025 & \bfseries 0.626\\
\bottomrule
\end{tabular}%
}
\end{table}

The matched comparison uses 60 frames, 20 per state, without temporal filtering. Every method receives the same frame and claim, with optional procedure text for Qwen3-VL. The verifier identifies 11 of 20 contradicted claims, compared with none for either Qwen3-VL protocol, demonstrating explicit invalid-state recognition alongside lower false support.

\noindent\textbf{Detector Protocol.} Table~\ref{tab:detector} uses 304 corrected images and 905 boxes at confidence 0.10 and intersection over union (IoU) 0.50. The unpaired synthetic baseline applies the detection objective to composed images without pair metadata. Box recall measures localization; correct recall also requires identity. Matched class accuracy conditions on localized matches. Correct F1 is the harmonic mean of correct recall and precision over all detections. Hard-pair error (HP err.) divides confusions by all ground-truth boxes, so localization can expose identity errors hidden by misses.

\begin{table}[t]
\centering
\caption{YOLO26s~\cite{jocher2026yolo26} calibration at confidence 0.10 and intersection over union (IoU) 0.50. Matched Acc. is class accuracy on localized boxes; hard-pair error (HP Err.) is normalized by all ground-truth boxes.}
\label{tab:detector}
\setlength{\tabcolsep}{0.35em}
\resizebox{\ifdim\width>\columnwidth\columnwidth\else\width\fi}{!}{%
\begin{tabular}{@{}l *{5}{S[table-format=1.3]}@{}}
\toprule
Training & {Box R. $\uparrow$} & {Correct R. $\uparrow$} & {Matched Acc. $\uparrow$} & {HP Err. $\downarrow$} & {Correct F1 $\uparrow$}\\
\midrule
Real images & 0.169 & 0.097 & 0.575 & 0.041 & 0.110\\
\addlinespace[0.25em]
Unpaired synthetic & 0.181 & 0.097 & 0.537 & 0.046 & 0.126\\
\addlinespace[0.25em]
PI-TwinSwap & \bfseries 0.459 & \bfseries 0.210 & 0.458 & 0.115 & \bfseries 0.190\\
\bottomrule
\end{tabular}%
}
\end{table}

PI-TwinSwap increases correctly localized and identified objects from 88 to 190 and improves Correct F1 beyond unpaired synthesis, expanding the evidence available for verification. The matched population also changes as localization improves. We therefore read matched identity accuracy together with coverage and hard-pair errors in Table~\ref{tab:detector}. All downstream evaluations keep this detector fixed.

\section{Conclusion}

We introduced \methodname{} to learn robot view selection from egocentric assembly assistance. The method separates evidence requirements from observation changes a robot can reproduce, then adapts view preferences through object-centered calibration. With candidate images hidden and no target-domain view labels for policy training, replay evaluations show higher human-rated verifiability on gearbox assemblies and a higher correct decision rate for the relative-view selector on IMPACT.

\section*{Acknowledgments}

The project is funded by the Deutsche Forschungsgemeinschaft (DFG, German Research Foundation) -- SFB-1574 -- 471687386. This work was supported in part by the SmartAge project sponsored by the Carl Zeiss Stiftung (P2019-01-003; 2021--2026). The authors gratefully acknowledge the computing time provided on the high-performance computer HoreKa by the National High-Performance Computing Center at KIT (NHR@KIT). This center is jointly supported by the Federal Ministry of Education and Research and the Ministry of Science, Research and the Arts of Baden-W\"urttemberg, as part of the National High-Performance Computing (NHR) joint funding program (\url{https://www.nhr-verein.de/en/our-partners}). HoreKa is partly funded by the German Research Foundation (DFG).

\bibliographystyle{IEEEtran}
\IEEEtriggeratref{43}
\bibliography{main}
\end{document}